\documentclass{article}

\usepackage[final]{cpal_2025}

\usepackage{url}
\usepackage{algorithm}
\usepackage{algorithmic}
\usepackage[nolist]{acronym}
\usepackage{xcolor}   
\usepackage{booktabs} 
\usepackage{amsmath}
\usepackage{mathtools}
\usepackage{multirow}
\usepackage{amsfonts}
\usepackage{todonotes}
\usepackage{subcaption}
\usepackage{tablefootnote}
\usepackage{array}
\usepackage{pdflscape}
\usepackage{graphics}
\usepackage{xspace}
\usepackage{caption}

\DeclareMathOperator{\quant}{Q}
\DeclareMathOperator{\dec}{D}
\DeclareMathOperator{\enc}{E}
\DeclareMathOperator{\bwmath}{bw}
\DeclareMathOperator{\softmath}{s}

\newcommand{\bw}{^{\bwmath}}
\newcommand{\soft}{^{\softmath}}
\newcommand{\bwsoft}{^{\bwmath,\softmath}}
\newcommand{\nbits}{n}
\newcommand{\nthr}{M}
\newcommand{\nfeat}{K}
\newcommand{\indfeat}{k}
\newcommand{\indthr}{m}
\newcommand{\datalen}{N}
\newcommand{\mytodo}[1]

\newcommand{\transpose}{\mathsf{T}}

\newcommand{\cpulong}{CPU Activity Database}
\newcommand{\cpu}{cpu\_act}
\newcommand{\sulfur}{sulfur}
\newcommand{\sulfurlong}{Sulfur}
\newcommand{\wine}{wine}
\newcommand{\winelong}{Wine Quality}
\newcommand{\california}{california}
\newcommand{\californialong}{California Housing}
\newcommand{\fried}{fried}
\newcommand{\friedlong}{Fried}
\newcommand{\superconduct}{superconduct}
\newcommand{\superconductlong}{Superconduct}

\title{ Trainable Bitwise Soft Quantization \\ for Input Feature Compression}

\author{%
  Karsten Schrödter\textsuperscript{1}, \hspace{0.2cm} Jan Stenkamp\textsuperscript{1}, \hspace{0.2cm} Nina Herrmann\textsuperscript{1}, \hspace{0.2cm} Fabian Gieseke\textsuperscript{1,2}
  \\
  \textsuperscript{1}University of Münster (Germany), \hspace{0.2cm}~\textsuperscript{2}University of Copenhagen (Denmark)\\
  \texttt{[first\_name].[last\_name]@uni-muenster.de}\\
  \texttt{fabian.gieseke@di.ku.dk}
}

\begin{document}

\begin{acronym}
\acro{DNN}[DNN]{Deep Neural Network}
\acro{IoT}[IoT]{Internet of Things}
\acro{AI}[AI]{Artificial Intelligence}

\end{acronym}

\maketitle

\begin{abstract}
The growing demand for machine learning applications in the context of the Internet of Things calls for new approaches to optimize the use of limited compute and memory resources.
Despite significant progress that has been made w.r.t. reducing model sizes and improving efficiency, many applications still require remote servers to provide the required resources.
However, such approaches rely on transmitting data from edge devices to remote servers, which may not always be feasible due to bandwidth, latency, or energy constraints.
We propose a task-specific, trainable feature quantization layer that compresses the input features of a neural network. This can significantly reduce the amount of data that needs to be transferred from the device to a remote server.
In particular, the layer allows each input feature to be quantized to a user-defined number of bits, enabling a simple on-device compression at the time of data collection. 
The layer is designed to approximate step functions with sigmoids, enabling trainable quantization thresholds. 
By concatenating outputs from multiple sigmoids, introduced as bitwise soft quantization, it achieves trainable quantized values when integrated with a neural network.
We compare our method to full-precision inference as well as to several quantization baselines.
Experiments show that our approach outperforms standard quantization methods, while maintaining accuracy levels close to those of full-precision models.
In particular, depending on the dataset, compression factors of $5\times$ to $16\times$ can be achieved compared to $32$-bit input without significant performance loss. 
\end{abstract}

\section{Introduction}
\label{sec:intro}
The number of \ac{IoT} applications has steadily grown in recent years, with use cases spanning a wide range of domains~\citep{rizvi_revolutionizing_2024,s20113113,smartcities4020024,HAGHIKASHANI2021103164}. These applications are often based on microcontrollers that are equipped with sensors that measure parameters such as temperature, humidity, pressure, or vibrations. Typically, such microcontrollers have very limited compute and memory capabilities (e.g., only 2\,KB of RAM), which severely restricts the types of algorithms that can be executed ``locally'' on the devices. This necessitates the development of efficient, resource-aware implementations tailored to the specific needs of the given application and hardware.

Several approaches have been proposed for deploying machine learning models in resource-constrained settings. One common strategy is to reduce the model size to an extent that inference can be performed on the device. Nevertheless, this approach requires trade-offs between model complexity and the available resources~\cite{WardenSitunayake2019,abs-2106-04008}.
Alternatively, the collected sensor data can also be transmitted to more powerful remote machines for further processing. However, this strategy also poses significant challenges. Reliable communication links and sufficient energy supplies might not be available. In extreme cases, devices are battery-powered and can only transmit very limited amounts of data---sometimes just a few bytes per hour. This is the case, for instance, with communication protocols such as LoRaWAN. Transmitting data via mobile networks (e.g., LTE) is in contrast often energy-intensive and might also not be possible due to a lack of cellular coverage~\citep{krizanovic_advanced_2023}.

We focus on scenarios where (a) on-device execution of a machine learning model is not feasible, and (b) only very limited amounts of data can be transmitted to a remote server. 
Such situations commonly arise when \ac{IoT} devices are deployed in remote locations, e.g., for environmental monitoring (forestry, wildlife, agriculture, \ldots). Hence, the devices mainly collect data, and strategies are required to minimize the size of data transmitted. Straightforward approaches are based on selecting important features (e.g., via forward feature selection~\cite{HastieTF2009}). Another strategy involves compressing the collected (sensor) data by using lower precision for floating-point numbers. However, naive precision reduction is task-agnostic and often degrades downstream model performance. 

\begin{figure*}[t]
    \centering
    \begin{subfigure}[b]{0.32\textwidth}
        \includegraphics[width=\linewidth]{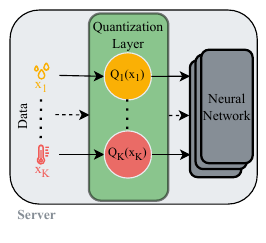}
        \caption{Training}
        \label{fig:network-split-training}
    \end{subfigure}\hspace{0.04\textwidth}
    \begin{subfigure}[b]{0.56\textwidth}
        \includegraphics[width=\linewidth]{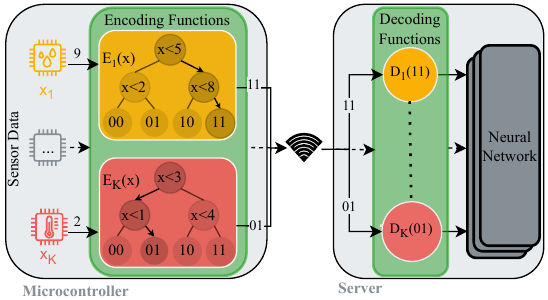}
                \caption{Inference}
                \label{fig:network-split-inference}
    \end{subfigure}
    \vskip-0.2cm
    \caption{
    The trainable feature quantization layer is integrated into a neural network to learn task-specific compressions for each input feature:
    (a) During training, the quantization layer (green rectangle) and the neural network are trained jointly on a remote server. The layer gives rise to an encoder-decoder
    composition $\quant_i= \dec_i \circ \enc_i$
    for the $i$-th input feature. (b) During inference, the encoder $\enc_i$ is used to encode the $i$-th feature on the resource-constrained device using lightweight coding logic. 
    The compressed features are then sent to the server, where the decoder $\dec_i$ is used to decode the $i$-th feature. All decoded features are then used as input for the remaining neural network, which is executed on the remote server. }
    \vskip-0.2cm
    \label{fig:network-split}
\end{figure*}

We propose an end-to-end trainable feature quantization layer that can be integrated into a (deep) neural network to learn task-specific compressions for each input feature. The resulting encoding scheme is based on simple intervals and can be efficiently implemented using standard \emph{if-then-else} rules.
As shown in Figure~\ref{fig:network-split}, the quantization layer is jointly trained with a given neural network on a remote server to determine optimal thresholds and quantized values for efficient deployment.
During inference, quantization is implemented on a resource-constrained device using lightweight logic, while decoding is performed on a more capable server, where the remaining neural network processes the compressed input. By integrating quantization into training, the method adapts to the underlying data distribution,
enabling flexible and effective splitting of input features.

\section{Background}
We start by summarizing related work and our contributions.
\subsection{Related Work}

Several approaches have been proposed to execute machine learning methods on \ac{IoT} nodes.  
The main challenge in this context is to reduce both computational and memory demands to enable execution on highly resource-constrained hardware---without significant loss in model accuracy. This has led to the \emph{tiny machine learning} paradigm~\cite{pmlr-v97-gural19a, abs-2106-04008, bonsai_Kumar_2017,WardenSitunayake2019}. When sufficiently optimized, such ``tiny'' models can run directly on microcontrollers, enabling compact and intelligent devices. However, reducing the model size generally leads to inferior results, especially when the \ac{IoT} devices are extremely resource-constrained. As a result, offloading the collected data to more powerful remote machines still remains a common approach for many applications.
Data transfer during the inference phase is also an active area of research. Broadly, existing approaches fall into two categories: (a) feature extraction, which transforms the original features into a new representation, and (b) feature selection, which identifies a relevant subset of the original features (e.g., \cite{pmlr-v97-balin19a,NIPS2016_6250,OehmckeG2022InputSelection,lossyAE,xu2018scaling,lemhadri2021lassonet}).

The research area of \textit{collaborative intelligence}~\citep{collabintelligence} explores the possibility of distributing inference tasks across multiple devices, with each device being responsible for computing a portion of the entire model. 
\citet{neurosurgeon_Kang2017} first proposed a collaborative intelligence approach, called \emph{Neurosurgeon}, which automatically determines split points for deep neural networks in a layer-wise manner.
Works based on this approach also consider reducing the size of the data to be transmitted by focusing on image compression \cite{bottlefit_Matsubara2022b, Matsubara2022, Singh2020, Yuan2022}.
Another topic of interest is to actually find suitable split points for neural networks~\citep{autosplit_BanitalebiDehkordi2021} and to identify so-called early-exit points~\citep{spinn_Laskaridis2020}. 

\textit{Quantization} is used to reduce the bit width of activations and weights of a neural network, see \citet{Gholami2021} or \citet{Li2024} for an overview. 
When it comes to training and inference, the two main approaches are based on applying quantization only during inference (post-training quantization) or on incorporating quantization into the training process (quantization-aware training). Research on post-training quantization includes, for example, output reconstruction~\cite{Zhao19,Nagel20} and clustering of activations~\cite{Cohen2020}. In quantization-aware training, non-differentiable functions are classically approximated by identity functions using the Straight-Through Estimator~(STE)~\cite{Bengio13}. 
Further research approximates non-differentiable quantization layers with random noise~\cite{AB2019} or applies differentiable soft quantization functions using a tanh function~\cite{softquant_Gong2019}. Furthermore, learnable quantizers have been investigated such as quantization with trainable step size \cite{Esser_2020} and learnable lookup tables \cite{Wan_2021_NBDT}. \citet{Yang2019} propose quantization networks and introduce soft quantization using sigmoid functions. Although this quantization enables flexible quantizer learning, empirical results show that learned thresholds generally do not outperform predefined ones.

\subsection{Contribution}
We focus on efficient quantization methods for input features. 
Building on the concept of soft quantization~\cite{Yang2019}, which performs layer-wise compression of activations and weights using predefined quantized values, we propose two modifications: first, we transform this approach into feature-wise compression of inputs, allowing for learnable thresholds per feature; second, we incorporate bitwise quantization to facilitate the learning of quantized values.
We also present a comprehensive evaluation of input data quantization across six datasets. Overall, our framework enables efficient feature quantization on low-power devices, requiring only a few if-else statements for encoding.

\section{Methodology}
\label{sec:methods}

\begin{figure*}[t]
    \centering
    \includegraphics[width=0.99 \linewidth]{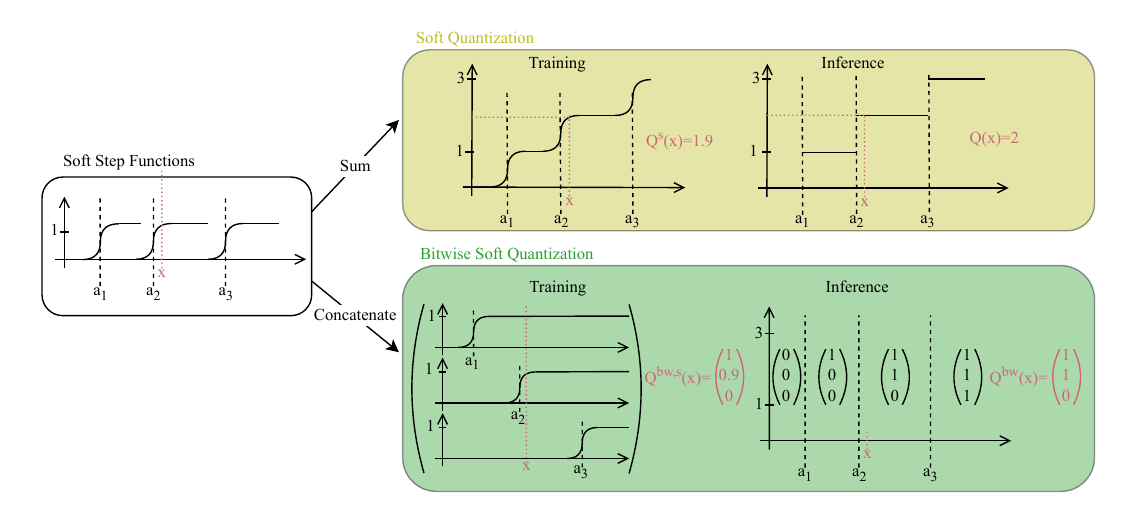}
    \vskip-0.2cm
    \caption{
    Schematic overview:
    Soft Quantization ($\quant\soft$) and Bitwise Soft Quantization ($\quant\bwsoft$) are formed by summing or concatenating multiple soft step functions. Both are differentiable with respect to thresholds, enabling optimization during training. During inference, they are converted into Hard Quantization ($\quant$) and Bitwise Quantization ($\quant\bw$) via rounding. 
    }
    \label{fig:stepfunction-quants}
\end{figure*}

A \emph{quantization function} $\quant: \mathbb{R} \to \{v_0,\dots, v_{\nthr}\}$ is based on $\nthr+1$ quantized values for some predefined $\nthr \in \mathbb{N}$. In this work, we consider scalars or binary vectors, i.e., $v_{\indthr} \in \mathbb{R}$ or $v_{\indthr} \in \{0,1\}^\nthr$ for $0 \le \indthr \le \nthr$.
The function $\quant$ can be split into an encoding function $\enc: \mathbb{R} \to \{0, \dots,\nthr\}$ and a decoding function $\dec: \{0,\dots,\nthr\} \to \{v_0,\dots,v_{\nthr}\}$, defined by $\indthr \mapsto v_{\indthr}$, so that $\quant = \dec \circ \enc$.\footnote{Note that $\quant$ and $\enc$ are often both called quantization functions. Since we may set the decoding function $\dec$ to the identity, each encoding function is also a quantization function.} In practice, a bit width of $\nbits$-bits for some given ${\nbits} \in \mathbb{N}$ is achieved by using $\nthr = 2^{\nbits}-1$.

In \emph{split inference} approaches, the data is first encoded using~$\enc$, resulting in $\nbits$-bit values that are sent to a remote server. There, the values are decoded using~$\dec$ and processed further. This process is illustrated in Figure~\ref{fig:network-split-inference}.

\subsection{Encoding Function via Thresholds}
\label{subsec:hard_quantization}
We introduce encoding functions that are based on multiple thresholds by combining step-functions via summation. For a threshold $a \in \mathbb{R}$, the hard step-function $\mathbb{I}_{\ge a}: \mathbb{R} \to \{0,1\}$ is defined as
$$
    \mathbb{I}_{\ge a}(x) = \begin{cases}
    1 & \mbox{if } x \ge a \\
    0 & \mbox{if } x < a
\end{cases}~.
$$
Multiple step-functions can be combined in order to define an encoding function based on thresholds as follows: consider $\nthr$ thresholds $a_1 < \dots < a_{\nthr} \in \mathbb{R}$. Then, the associated hard encoding function $\enc_{a_1,\dots,a_{\nthr}}: \mathbb{R} \to \{0, 1, \dots, {\nthr}\}$ can be defined as 
\begin{align}
x &\mapsto \sum\limits_{\indthr=1}^{\nthr} \mathbb{I}_{\ge a_{\indthr}}(x) \in \{0,\dots,{\nthr}\}\text{.}
\end{align}

\subsection{Decoding Functions}

In order to define a suitable decoding function, quantized values need to be defined. The trivial approach is to define the decoding function as the identity, so that $v_\indthr=\indthr$ for all $0 \le \indthr \le \nthr$, see Hard Quantization in Figure \ref{fig:stepfunction-quants} for an example. We present two more ideas.

\begin{figure*}[t!]
    \centering
    \includegraphics[width=\linewidth]{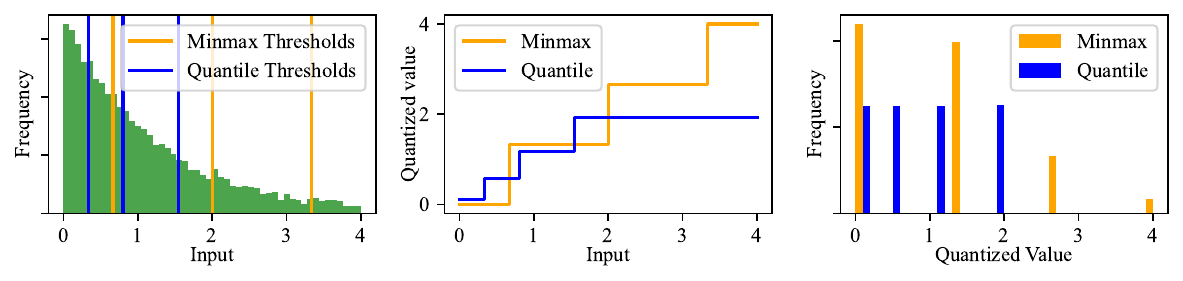}
    \vskip-0.1cm
    \caption{Example of a quantization with bit width $\nbits=2$ (i.e., $M=3$ thresholds) using minmax and quantile quantization on artificial skewed data. Left: Histogram of input data together with thresholds for minmax and quantile quantization. Center: Minmax and quantile quantization functions. Right: Histogram of quantized values for minmax and quantile quantization.}
    \label{fig:minmax_and_quantile_quantization}
\end{figure*}

Another option is to resort to quantized values in the middle of the interval between two thresholds. For $\nthr > 1$ and $0 \le \indthr \le \nthr$ we define
\begin{equation}
    \centering
    \label{eq:quantized_values}
        v_{\indthr} \coloneq \frac{1}{2} \cdot (a_{\indthr} + a_{\indthr+1})\text{.}
\end{equation}
We need to define two auxiliary thresholds $a_0 \coloneq  2\cdot a_1 - a_2$ and $a_{\nthr+1} \coloneq  2\cdot a_\nthr - a_{\nthr-1}$. 
The decoding function is then given by $\dec_{a_1,\dots,a_{\nthr}}: \{0,\dots,{\nthr}\} \to \{v_0,\dots,v_{\nthr}\}$ with $\indthr \mapsto v_\indthr$. Finally, the quantization function associated with the thresholds $a_1 < \dots < a_{\nthr} \in \mathbb{R}$ is then given by 
$$ \quant_{a_1,\dots,a_{\nthr}} = \dec_{a_1,\dots,a_{\nthr}} \circ \enc_{a_1,\dots,a_{\nthr}}. $$
For instance, the \emph{minmax quantization} divides the range of data points into equally sized intervals, where quantized values are given by the middle of the intervals, see, e.g., \citet[p. 5]{Gholami2021}. 
The idea of \emph{quantile quantization} is to define the thresholds using quantiles, such that all quantized values occur equally often. Using quantile thresholds in the quantization function defined above leads to a slight modification of quantile quantization as described by \citet[App. F.2]{Dettmers2021}. We refer to Figure~\ref{fig:minmax_and_quantile_quantization} for a visualization and to Appendix~\ref{ss:minmax_and_quantile_quantization} for a detailed description of both minmax and quantile quantization.

\subsubsection{Bitwise Quantization via Thresholds}
Another option to define decoding functions is to consider binary vectors as quantized values. The underlying idea is to have a quantization function that concatenates step-functions, instead of summing them.  
For $0 \le \indthr \le \nthr$, we define the bitwise quantized value $v_\indthr\bw$ as follows:
\begin{equation}
\label{eq:quant_vals_bitwise}
v_{\indthr}\bw \coloneq [
    \underbrace{1, 1, \ldots, 1}_{m \text{ ones}}, \underbrace{0, 0, \ldots, 0}_{M-m \text{ zeros}}
]^\transpose \in \{0,1\}^{\nthr}.
\end{equation}
The decoding function $\dec\bw: \{0,\dots,\nthr \} \to \{0,1\}^{\nthr}$ with $\indthr \mapsto v_{\indthr}\bw$ is the bitwise decoding function and the composition $\quant\bw_{a_1,\dots,a_{\nthr}} = \dec\bw \circ \enc_{a_1,\dots,a_{\nthr}}: \mathbb{R} \to \{0,1\}^{\nthr}$ is called bitwise quantization, where $a_1 < \dots < a_{\nthr} \in \mathbb{R}$ are $\nthr$ thresholds and $\enc_{a_1,\dots,a_{\nthr}}$ is the encoding function defined above. For $x \in \mathbb{R}$ the bitwise quantization is given by 
\begin{equation}
	\quant\bw_{a_1,\dots,a_{\nthr}}(x) = [\mathbb{I}_{\ge a_{1}}(x),\dots,\mathbb{I}_{\ge a_{\nthr}}(x)]^T \in \{0,1\}^\nthr. 
\end{equation}
See Bitwise Soft Quantization in Figure~\ref{fig:stepfunction-quants} for an example.

Using bitwise quantized values as input for a neural network can be helpful, as composing $\quant \bw_{a_1,\dots,a_{\nthr}}$ with a linear layer $h: \mathbb{R}^{\nthr} \to \mathbb{R}$ yields a quantization function $h \circ \quant \bw_{a_1,\dots,a_{\nthr}}: \mathbb{R} \to \{\tilde{v}_0,\dots, \tilde{v}_{\nthr}\}$ with learnable quantized values. We refer to Appendix \ref{app:trainable_quantized_values} for mathematical details.

In the context of multiple inputs, a linear layer that receives bitwise quantized values as input and produces a single neuron can be viewed as mapping each bitwise-quantized feature to a learned quantized scalar and then summing these scalars to produce the output.
A similar layer with multiple outputs is therefore able to process different quantized values per input feature.
Thus, when applying bitwise quantization to each input feature and integrating it with a neural network, the first layer of the network effectively learns and combines quantized values in diverse ways tailored to the specific task. This approach can be beneficial, as optimal quantized values may vary depending on the task.

\subsection{Approximating with Soft Quantization}

This section introduces soft quantization, which enables the training of thresholds using gradient-based methods. Soft quantization functions, introduced by \citeauthor{Yang2019}, are approximations of hard quantization functions, where the step-functions are replaced by sigmoidal functions to avoid vanishing gradients. Building on this, we also introduce bitwise soft quantization.

\subsubsection*{Soft Step Function}

Soft step functions are translated and stretched sigmoid functions. For $a \in \mathbb{R}$ and a temperature parameter $\tau > 0$ the soft step function $\mathbb{I}_{\ge a}\soft: \mathbb{R} \to (0,1)$ is 
\begin{equation}
	\mathbb{I}\soft_{\ge a}(x) = \frac{1}{1+ \exp\left(\frac{a-x}{\tau}\right)} = \sigma \left(\frac{x-a}{\tau}\right)\text{,}
\end{equation}
where $\sigma: \mathbb{R} \to \mathbb{R}$ is the sigmoid function. Note that the threshold $a$ is used to translate the sigmoid function and $\tau$ is a stretching factor, inspired by the temperature parameter in distilling \cite{Hinton15}. Due to the shape of the sigmoid function, when rounding the output of the soft step function, the step function is reproduced, i.e.
$$ \mathbb{I}_{\ge a_{\indthr}}(x) = \mbox{round}\left(\mathbb{I}\soft_{\ge a_{\indthr}}(x) \right) ~ \forall x \in \mathbb{R}\text{.} $$

\subsubsection*{Soft Encoding and Quantization Function}
As with hard quantization functions, soft quantization functions are built by combining multiple soft step functions either through summation (soft quantization) or concatenation (bitwise soft quantization). 
As above, consider $\nthr$ thresholds $a_1  < \dots < a_\nthr \in \mathbb{R}$ and additionally a temperature parameter $\tau > 0$. Define the associated soft quantization function $\quant\soft_{a_1,\dots,a_\nthr}: \mathbb{R} \to (0, \nthr)$ and bitwise soft quantization function ${\quant}\bwsoft_{a_1,\dots,a_\nthr}: \mathbb{R} \to (0, 1)^\nthr$ by 
\begin{align}
\quant\soft_{a_1,\dots,a_\nthr}(x) &=\sum\limits_{i=1}^\nthr \mathbb{I}^s_{\ge a_{\indthr}}(x) \in (0,\nthr) \\
\quant\bwsoft_{a_1,\dots,a_\nthr}(x) &=[\mathbb{I}^s_{\ge a_{1}}(x),\dots,\mathbb{I}^s_{\ge a_{\nthr}}(x)]^T \in (0,1)^\nthr 
\end{align}
for $x \in \mathbb{R}$. See (Bitwise) Soft Quantization in Figure \ref{fig:stepfunction-quants} for an example. The notation for soft quantization slightly differs from the one in \citet{Yang2019}, but the resulting functions have the same shape. The approximation of the (bitwise) hard quantization function by the (bitwise) soft quantization function improves with decreasing $\tau > 0$. Note that (bitwise) soft quantization is differentiable with respect to all thresholds $a_{\indthr}$. 
For more details on the derivative, we refer to Appendix \ref{app:softencoding}.
The (bitwise) soft quantization function can be used to learn thresholds using gradient-based methods. 

\subsection{Quantization Layer}

A quantization layer quantizes each input feature independently and concatenates the quantized values into one output vector. We refer to Appendix \ref{app:quantization_layer} for a mathematical description. If all quantization functions are soft quantization functions, the layer is called a soft quantization layer. Included at any step of a deep neural network, a soft quantization layer can be used in order to learn the best quantization thresholds. Note that during inference, all soft quantization functions are converted to hard quantization functions to ensure a smaller bit width for the encoded values. Therefore, it makes sense to decrease $\tau$ during the training process in order to improve the approximation of a hard quantization layer and decrease the effect of rounding.

\section{Experiments}
We compare the performance of soft bitwise quantization with that of several baseline quantization methods on a number of regression datasets. Additionally, we conduct an ablation study to differentiate between soft and bitwise quantization. 
To demonstrate the minimal overhead caused by our encoding approach, we conducted additional experiments on the latency and energy requirements of an implementation on a microcontroller. The respective results are included in Appendix~\ref{app:mcu_deployment}. \footnote{The code for all experiments is available at \url{https://github.com/kschr40/FeatureCompression}}

\subsection{Experimental Setup}

The following sections describe the evaluated datasets, models, and training process.

\subsubsection{Datasets}
As features are represented with a varying number of bits, our approach is tested on regression tasks using $6$ datasets that have continuous features with a variety of values. The prediction tasks of the datasets cover a wide range; from housing prices (dataset name \californialong  ~\citep{pace1997sparse}) or relative compute time of cpu in user mode (\cpulong \footnote{\url{https://www.openml.org/search?type=data&status=active&id=197}}), sinusoidal formula reproduction (\friedlong  ~\citep{friedman1991multivariate, breiman1996bagging}), hydrogen sulfide concentration (\sulfurlong ~\citep{fortuna2007soft}), temperature of superconductors (\superconductlong ~\citep{superconductoriginal}) and quality of wine (\winelong ~\citep{wine_qualitydata}). The number of instances varies from about 6500 in \winelong ~to more than 40000 in \friedlong, while the datasets have between 7 (\sulfurlong) and 80 (\superconductlong) features. The \winelong ~dataset also has on average the fewest distinct values per feature, about 241. In contrast, the other datasets contain on average more than 1000 distinct values per feature, up to 8878 for \californialong. We refer to Appendix \ref{app:datasets}, Table \ref{tab:datasets} for more information about the datasets.  

\begin{table}[t]
  \centering
  \begin{tabular}{lll}
    \toprule
    Name & Abbreviation  & Source \\ \midrule
    Full Precision     & FP &  Baseline \\ 
    Pre Minmax Quantization   & Pr-MQ   & Baseline \\
    Pre Quantile Quantization & Pr-QQ  & Baseline \\ 
    Learnable Step Size Quantization & LSQ & \cite{Esser_2020} \\
    Learnable Lookup Table with granule 4 & LLT4 & \cite{Wan_2021_NBDT} \\
    Learnable Lookup Table with granule 9 & LLT9 & \cite{Wan_2021_NBDT} \\
    Bitwise Soft Quantization & Bw-SQ   & Ours \\
    \bottomrule
  \end{tabular}
 \vspace{0.2cm}
  \caption{Models}
  \label{tab:models}
 \vspace{-0.2cm}
\end{table}

\subsubsection{Models} 

We propose to use a Bitwise Soft Quantization (Bw-SQ) model for joint training of quantization and the subsequent neural network. The thresholds for the bitwise soft quantization layer are initialized using the thresholds of quantile quantization. 

In order to benchmark our model, five approachess are analyzed as baselines, see Table \ref{tab:models} for an overview of used models. The first model is a plain multi-layer perceptron (MLP) using bit width $32$ without any quantization, called the full precision (FP) model. Two additional baselines are minmax and quantile quantization, applied to the input features during training and inference, simulating a quantization-aware training approach for input feature compression. Furthermore, we use two more recent approaches, described in more detail below. 

\textbf{Learned Step Size Quantization \cite{Esser_2020} (LSQ)}
This approach approximates the gradient of the quantization function by using a straight-through estimator \cite{Bengio13}, which effectively leads to a learnable step size (or scale parameter) in quantization. Due to lack of an official github repository, we use an unofficial one\footnote{https://github.com/zhutmost/lsq-net} and include the code in our training pipeline. 

\textbf{Learnable Lookup Tables \cite{Wan_2021_NBDT} (LLT)} 
This work formulates quantization as a lookup table, made differentiable by approximating a hard decision using a softmax operation. In consequence, learnable lookup tables start by creating too many thresholds (by a factor called granule) and learns to keep the best thresholds. In the original work, a granule of $9$ was suggested. In contrast, preliminary experiments suggested that granule $4$ works well for our use cases. Therefore, we are comparing against both variants in our experiments. 

\begin{figure}[t]
	\centering
	\includegraphics[width=\linewidth, trim={2cm 0 2cm 0},clip]{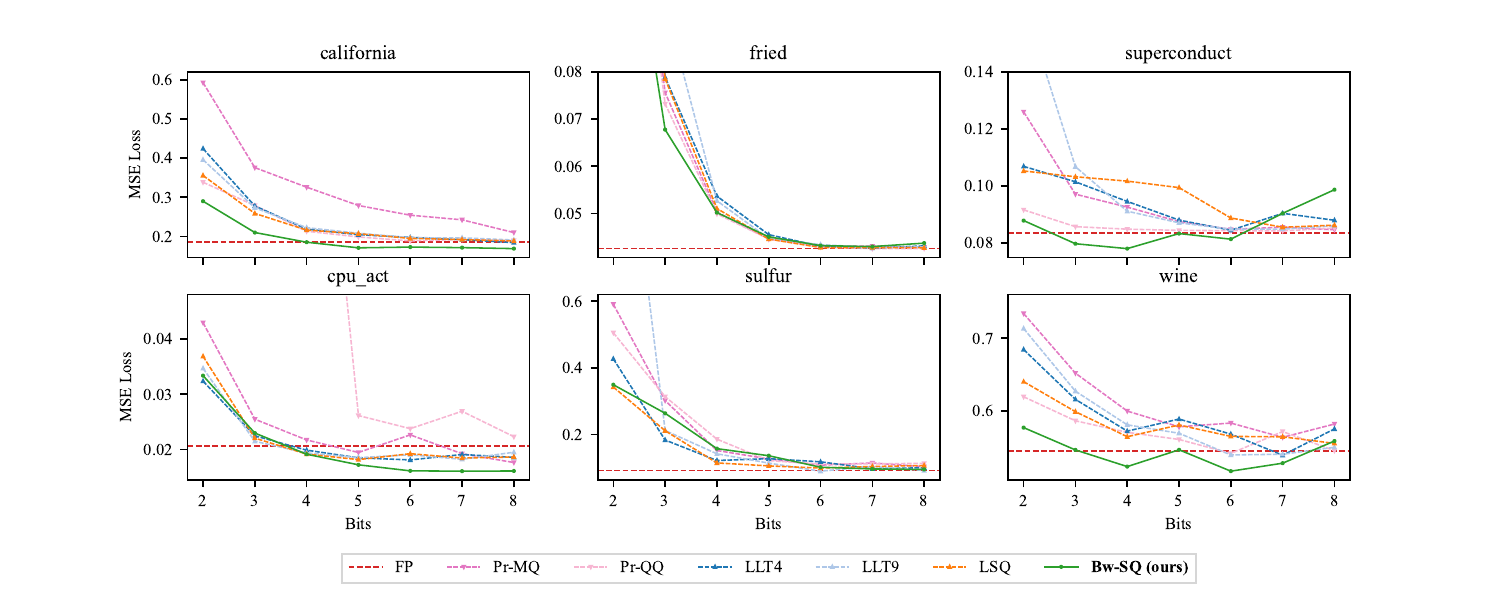}
	\vskip-0.1cm
	\caption{
		MSE values per dataset per quantization level of the best-performing hyperparameter setting averaged over 10 data splits for all methods. The red line is the average over the full precision measurements.
	}
	\label{fig:linplotdatasets}
\end{figure}

\subsubsection{Model Training and Selection}
In order to compare the performance of our models with the baseline models, hyperparameter tuning is performed for all methods. This is done to ensure a fair comparison between all methods, which might perform best using different hyperparameter settings. The basic architecture for all models (except the FP model) contains a quantization layer and a subsequent MLP. The quantization layer compresses each feature to the desired bit width. In our experiments, the bit width varies from $\nbits = 2 $ bits to $\nbits = 8$ bits per feature. The MLP is a dense neural network, where the number of hidden layers and neurons per hidden layer are hyperparameters. 
All models are trained using mean squared error (MSE) loss together with dropout. The dropout rate, the learning rate, and the number of epochs to train are also hyperparameters. An additional hyperparameter is given by the decrease factor of the temperature parameter $\tau$ for soft quantization layers. $\tau$ is initialized as $1$ and exponentially decreased after each epoch, such that it equals the decrease factor at the end of training, see also Appendix \ref{app:tau_schedule} for justification of this choice. During preprocessing, all features and labels are standardized. 
$4$-fold cross-validation is used to determine the best hyperparameter setup per method, bit width and dataset. 
Finally, the best hyperparameter setting is applied with $10$ different train-test splits ($90\%$ train, $10\%$ test data) to get a robust estimate of the generalization performance. In addition to the average MSE of the best performing hyperparameter setup, a $95\%$-confidence interval for the average MSE across all $10$ data splits is calculated. The results of two methods are called significantly different if the confidence intervals are disjoint. 
For details on hyperparameter tuning, see Appendix~\ref{app:hyperparameter}.

\subsection{Results}
\label{sec:resultskfold}
The main results of our experiments are visualized in Figure~\ref{fig:linplotdatasets}, showing the average test loss of the trained MLPs for the different bit widths. The corresponding table of results, as well as a table of $95\%$-confidence intervals, are contained in Appendix \ref{app:result}. 
With a few exceptions, the loss is decreasing for all quantization methods with increasing bit width across the evaluated datasets, converging to each other and towards the performance of the FP model.
In over half of the experiments (26 out of 42), Bw-SQ yields the smallest average MSE loss across all quantization methods.

\textbf{Comparison with Quantization Techniques.} 
For the datasets \winelong, \californialong, \cpulong ~and \friedlong, it can be seen that Bw-SQ outperforms the comparison quantization methods, partly by far. On \superconductlong, Bw-SQ also performs well for small bit widths, but the error increases for higher bit widths, which might indicate numerical instability on this dataset. Finally, Bw-SQ only performs well on \sulfurlong ~dataset for bit widths $2$ and $6$ to $8$ and is outperformed by e.g. LSQ for small bit widths up to $6$. In particular, there is no quantization technique that performs best for all datasets. For the comparison methods, it can be seen that in general LLT works better with the smaller granule $4$ for small bit width. Between LLT and LSQ, the superior model is dataset-dependent. Another observation is that comparing pre-defined thresholds, Pr-QQ performs better on 3 datasets (\winelong, \californialong ~and \superconductlong), Pr-MQ performs better on 2 datasets (\cpulong ~and \sulfurlong), while the performance is similar on \friedlong. Therefore, quantile thresholds do not always outperform minmax thresholds. Nevertheless, Bw-SQ -- initialized with quantile thresholds -- is still able to perform well on \cpulong ~due to learnable thresholds.   

\textbf{Comparison With Full Precision Model.}
We investigate the bit width tipping point, defined as the minimal bit width, such that there is no significant difference between Bw-SQ and FP. This varies between the datasets. 
For \superconductlong ~and \winelong, there is no significant difference for a bit width of $\nbits=2$, having a small performance drop of $5\%$ from FP to Bw-SQ for \superconductlong ~and $6\%$ for \winelong. 
For \californialong ~and \cpulong, the tipping point is at bit width $\nbits=3$ and Bw-SQ even performs better than FP for bit widths up from $4$ bits.
On the datasets \sulfurlong, the tipping point for bit width equals $\nbits = 4$, although FP (MSE $0.093$) performs much better in average than Bw-SQ (MSE $0.158$). On the artificial dataset \friedlong, the tipping point is at bit width $\nbits = 6$. Overall, Bw-SQ is able to compress data by an average compression factor of $11.1\times$ with a range from $5\times$ to $16\times$ without a significant performance loss compared to the FP model with bit width $32$. 

\subsection{Ablation Study}

To test which component of Bitwise Soft Quantization is responsible for the performance gain and to what extent, we perform an ablation study. We evaluate three variants of our model: The first model performs Soft Quantization (SQ), i.e. it has trainable thresholds, while the second and third model use Bitwise Minmax Quantization (Bw-MQ), resp. Bitwise Quantile Quantization (Bw-QQ). 
In Table \ref{tab:ablation} we collect the relative performance to Bitwise Soft Quantization, averaged over the datasets. More detailed results are presented in Appendix \ref{app:result}, Table \ref{tab:ablation_full}. It can be seen that none of the ablation methods can robustly perform similarly to Bw-SQ. For example, Bw-QQ slighly outperforms Bw-SQ on \winelong ~and \superconductlong, but does not perform good on \sulfurlong ~and \cpulong. In average, SQ is the best of the ablation methods. Nevertheless, it performs approximately $12\%$ worse than Bw-SQ. In conclusion, it can be seen that Bw-SQ is able to join the advantage of learnable thresholds from SQ and of the bitwise decoding as in Bw-QQ. 

\begin{table}[H]
\centering
    \begin{tabular}{l|rrrrrr|c}
\toprule
Dataset & \california & \fried & \superconduct & \cpu & \sulfur & \wine & Mean \\
\midrule
Bw-SQ & $0.195$ & $0.061$ & $0.086$ & $0.020$ & $0.172$ & $0.543$ & $0.180$ \\
\midrule
SQ & $+12.15 \%$ & $+3.74 \%$ & $+5.86 \%$ & $+28.34 \%$ & $+13.36 \%$ & $+7.36 \%$ & $+11.80 \%$ \\
Bw-MQ & $+57.43 \%$ & $+8.03 \%$ & $+16.42 \%$ & $+5.74 \%$ & $+21.21 \%$ & $+4.86 \%$ & $+18.95 \%$ \\
Bw-QQ & $+9.91 \%$ & $+6.64 \%$ & $-0.75 \%$ & $+112.85 \%$ & $+33.49 \%$ & $-1.36 \%$ & $+26.80 \%$ \\
\bottomrule
\end{tabular}

    \vspace{0.2cm}
    \caption{Average MSE of Bitwise Soft Quantization per dataset together with average MSE ratio of ablation methods
    against Bitwise Soft Quantization per dataset. Values below $0$ indicate better performance than Bitwise Soft Quantization.}  
    \label{tab:ablation}
 \vspace{-0.2cm}
\end{table} 

\section{Conclusion}
\label{sec:conclusion}

In this work, we introduce Bitwise Soft Quantization as a framework to combine learnable thresholds from soft quantization \cite{Yang2019} with learnable quantized values from bitwise quantization. This leads to a quantization technique that can be used to compress input features of neural networks by a factor $5\times$ to $16\times$ without significant performance loss compared to the $32$ bit full precision model. Results indicate that the regularization effect of quantization can even lead to better generalization on some datasets. However, this work is limited by its focus solely on MLP models and regression data, as well as the equal bit width compression applied to all features. Addressing these limitations will be the focus of future research.

\newpage 
\section*{Acknowledgements}
This work was funded by the German Federal Ministry for the Environment, Nature Conservation, Nuclear Safety and Consumer protection, Project TinyAIoT, Funding Nr. 67KI32002A. We also acknowledge the computational resources provided by the PALMA II cluster at the University of Münster (subsidized by the DFG; INST 211/667-1).
\bibliography{references}


\appendix


\newpage

\section{Minmax and Quantile Quantization}
\label{ss:minmax_and_quantile_quantization}

\subsection{Minmax Quantization}

The \emph{minmax quantization} divides the range of data points into equally sized intervals, where quantized values are given by the middle of the intervals, see, e.g., \citet[p. 5]{Gholami2021}. It can be recovered from our notation of quantization as follows: 

For a given dataset of length $\datalen \in \mathbb{N}$ consider all values of a selected continuous feature by $x_1,\dots, x_{\datalen} \in \mathbb{R}$. Define the minimal and maximal value as $ x_{min} \coloneq \min \{x_1,\dots, x_{\datalen}\}$ and  $x_{max} \coloneq \max \{x_1,\dots, x_{\datalen}\}$. Let $\nbits \in \mathbb{N}$ be the targeted bit width, then define $\nthr \coloneq 2^{\nbits} -1$ and the scale parameter $s$ by 
$$ s \coloneq \frac{x_{max}-x_{min}}{\nthr}\text{.} $$
Setting the thresholds $a_1 < \dots < a_M \in \mathbb{R}$ to
$$ a_{\indthr} \coloneq x_{min} + (\indthr - \frac{1}{2}) \cdot s, ~ 1 \le \indthr \le M 
$$
leads to quantization values (compare Equation \ref{eq:quantized_values})
$$ v_{\indthr} = x_{min} + \indthr \cdot s, ~0 \le \indthr \le M  \text{.}
$$

The resulting quantization function can be reformulated as follows:

For $0 \le \indthr \le M-1$ consider some $x \in [a_{\indthr}, a_{\indthr+1})$. Then the following chain of inequalities hold true 
$$ x_{min} + (\indthr - \frac{1}{2}) \cdot s = a_{\indthr} \le x < a_{\indthr + 1} = x_{min} + (\indthr + \frac{1}{2}) \cdot s \text{.}$$
By subtracting $x_{min}$ and dividing by $s$, this is equivalent to 
$$ \indthr - \frac{1}{2} \le \frac{x-x_{min}}{s} < \indthr + \frac{1}{2} \text{.} $$
Thus, we may deduce that 
$$x \in [a_{\indthr}, a_{\indthr+1}) \Leftrightarrow \indthr = \mbox{round}(\frac{x-x_{min}}{s}).$$
Therefore, the associated quantization function $\quant_{a_1,\dots,a_{\nthr}}: \mathbb{R} \to \{v_1,\dots,v_M\} \subset \mathbb{R}$ is given by 
$$ \quant_{a_1,\dots,a_{\nthr}}(x) = x_{min} + s \cdot \mbox{round} \left( \frac{x - x_{min}}{s} \right)$$ 
and thus is a slight modification of the quantization in \citet[p. 5]{Gholami2021} using the minimum and maximum of the signal as clipping range. Therefore, we will refer to this quantization function as minmax quantization. 
Please note that minmax quantization is sensitive to outliers as it works with minimal and maximal values. 

\subsection{Quantile Quantization}

The idea of quantile quantization is to define the thresholds using quantiles, such that all quantized values occur equally often. 

For a given dataset of length $\datalen \in \mathbb{N}$ consider all values of a selected continuous feature by $x_1,\dots, x_{\datalen} \in \mathbb{R}$. For $\tau \in [0,1]$ denote the $\tau$-quantile of $x_1,\dots,x_{\datalen}$ by $q_\tau(x_1,\dots,x_{\datalen})$.  Let $\nbits \in \mathbb{N}$ be the targeted bit width, then define $\nthr \coloneq 2^{\nbits} - 1$ and set 
$$ a_{\indthr} \coloneq q_{\tau_{\indthr}}(x_1,\dots,x_{\datalen}) \mbox{ with } \tau_{\indthr} = \frac{\indthr}{\nthr+1}, ~ 1 \le \indthr \le \nthr \text{.}$$

Using this, the quantized values $v_1,\dots,v_{\nthr}$ are given by Equation \ref{eq:quantized_values}. The associated quantization function $\quant_{a_1,\dots,a_{\nthr}}: \mathbb{R} \to \{v_1,\dots,v_{\nthr}\}$ will be called the quantile quantization. Note that this is a slight modification of quantile quantization as in \citet[App. F.2]{Dettmers2021}. 
Please note that quantile quantization is less sensitive to outliers, as it works with quantiles instead of minmal and maximal values of a distribution.

\section{Trainable Quantized Values via Bitwise Quantization}
\label{app:trainable_quantized_values}
As in Section \textit{Bitwise Quantization via Thresholds}, consider the bitwise quantization function $\quant \bw_{a_1,\dots,a_{\nthr}}:\mathbb{R} \to \{0,1\}^{\nthr}$ and let $h: \mathbb{R}^{\nthr} \to \mathbb{R}$ be a linear layer with $\nthr$ input dimensions and $1$ output dimension, i.e., 
$h(z)=w^\transpose \cdot z + b$ for $z \in \mathbb{R}^{\nthr}$ and some $w \in \mathbb{R}^\nthr$ and $b \in \mathbb{R}$. 
Then, for $x \in [a_\indthr, a_{\indthr + 1})$, we have (compare Equation \ref{eq:quant_vals_bitwise}):
\begin{align}
 \label{eq:trainable_quant_values} 
    \tilde{v}_\indthr & \coloneq h \circ \quant\bw_{a_1,\dots,a_{\nthr}}(x) = h(v_{\indthr}\bw) = w^\transpose \cdot v_{\indthr}\bw + b \\
    &= \sum\limits_{j = 1}^{\indthr} w_j + b  \text{.}\nonumber
\end{align}
Thus, $h \circ \quant\bw_{a_1,\dots,a_{\nthr}}: \mathbb{R} \to \{\tilde{v}_0,\dots, \tilde{v}_{\nthr}\}$ is itself a quantization function with the same thresholds $a_1 < \dots < a_\nthr$, but with learnable quantized values $\tilde{v}_{\indthr}$ for $0 \le \indthr \le \nthr$.  

Note that one option would be to set the linear layer by $b \coloneq v_0$ and $w = [w_1, \dots, w_\nthr]^T$ with $w_\indthr \coloneq v_\indthr - v_{\indthr - 1}$ for $1 \le \indthr \le \nthr$. Using Equation \ref{eq:trainable_quant_values}, it follows 
\begin{align}
 \label{eq:trainable_quant_values} 
    \tilde{v}_\indthr &= \sum\limits_{j = 1}^{\indthr} w_j + b  \nonumber = \sum\limits_{j = 1}^{\indthr}  v_\indthr - v_{\indthr - 1} + v_0 = v_\indthr  \nonumber
\end{align}
for all $0 \le \indthr \le \nthr$. Therefore, the quantization function $\quant_{a_1,\dots,a_{\nthr}}$ is an example of $h \circ \quant\bw_{a_1,\dots,a_{\nthr}}$. Nevertheless, the composition $h \circ \quant\bw_{a_1,\dots,a_{\nthr}}$ is much more flexible. 

\section{Derivative of Soft Quantization Functions}
\label{app:softencoding}

The partial derivative of the soft quantization function with respect to the thresholds is given by 

\begin{align*}
&\frac{\partial \quant\soft_{a_1,\dots,a_M}}{\partial a_{\indthr}}(x) = \frac{\partial \mathbb{I}\soft_{\ge a_{\indthr}}}{\partial a_{\indthr}}(x) = \frac{\partial \sigma\left(\frac{x - a_{\indthr}}{\tau}\right)}{\partial a_{\indthr}} \\ &=  \frac{-1}{\tau} \cdot \sigma\left(\frac{x - a_{\indthr}}{\tau}\right) \cdot \left(1- \sigma\left(\frac{x - a_{\indthr}}{\tau}\right)\right) \\ & = \frac{-1}{\tau} \cdot \mathbb{I}\soft_{\ge a_{\indthr}}(x) \cdot (1-\mathbb{I}\soft_{\ge a_{\indthr}}(x))
\end{align*}
for all $1 \le \indthr \le \nthr$.
This follows from the chain rule and the fact that the derivative of the sigmoid function is given by $$\frac{d\sigma}{dx}(x) = \sigma(x) \cdot (1-\sigma(x))$$ for all $x \in \mathbb{R}$. Therefore, the partial derivative of the soft quantization function with respect to a threshold behaves itself as the derivative of a sigmoid function, i.e. it is smooth and non-vanishing but it tends to $0$ as $x$ moves away from the threshold.  

Similarly, since $(\quant\bwsoft_{a_1,\dots,a_M})_\indthr = \mathbb{I}\soft_{\ge a_{\indthr}}$, the partial derivative of the $i$-th component of the bitwise quantization with respect to $a_\indthr$ vanishes for $i \ne \indthr$, while 
$$ \frac{\partial (\quant\bwsoft_{a_1,\dots,a_M})_\indthr}{\partial a_{\indthr}}(x) = 
\frac{-1}{\tau} \cdot \mathbb{I}\soft_{\ge a_{\indthr}}(x) \cdot (1-\mathbb{I}\soft_{\ge a_{\indthr}}(x)) \text{.}
$$

\section{Notation for Quantization Layer}
\label{app:quantization_layer}

A quantization layer quantizes each input feature independently. This appendix introduces the notation for a quantization layer. Let $\nfeat \in \mathbb{N}$ be the number of features and consider for each feature $1 \le \indfeat \le \nfeat$ a quantization function 
$Q_{\indfeat}: \mathbb{R} \to \{v_0^{(\indfeat)},\dots,v_{M_i}^{(\indfeat)}\}$, where $\nthr_{\indfeat} \in \mathbb{N}$ is the number of thresholds for feature $\indfeat$ and $v_0^{(\indfeat)}, \dots, v_{\nthr_{\indfeat}}^{(\indfeat)}$ are the quantized values. Note that $v_\indthr^{(\indfeat)} \in \mathbb{R}$ or $v_{\indthr}^{(\indfeat)} \in \{0,1\}^{\nthr_{\indfeat}}$ for all $1 \le \indthr \le \nthr_{\indfeat}$.   
Then the function defined by the quantization layer is given by 
$$ Q = (Q_1, \dots, Q_{\nfeat}): \mathbb{R}^{\nfeat} \to \prod\limits_{\indfeat=1}^\nfeat \{v_1^{(\indfeat)},\dots,v_{\nthr_{\indfeat}}^{(\indfeat)}\} \subset \mathbb{R}^{N'}$$
If all quantization functions are soft quantization functions, the layer is called a soft quantization layer. Note that since soft quantization is differentiable with respect to the thresholds, so is the soft quantization layer. In this case, the number of features, the number of thresholds and the temperature parameters are hyperparameters of the layer, while the thresholds are the parameters that can be learned.

\section{Deployment Experiments}
\label{app:mcu_deployment}
For the evaluation of energy and latency overhead we deployed our approach on a microcontroller and measured processing time and energy consumption. We utlized a \texttt{Walter} Internet of Things (IoT) system-on-module\footnote{\url{https://www.quickspot.io}} equipped with an ESP32-S3 CPU. The ESP32-S3 is used in multiple microcontrollers, and the code can easily be transferred to other hardware. We use the Power Profiler Kit II for power profiling. Measurements were taken at a static voltage supply of 3,700 mV, which is commonly used in lithium-ion batteries.

\subsection{Encoding}

We implemented an exemplary encoder for bit widths of 2, 3, and 4 and deployed it on the microcontroller. As shown in Table~\ref{tab:encoding_deployment} the latency experiments show that encoding the input features takes only microseconds and consumes only microjoule, which are negligible compared to the time it takes to measure or transmit the respective data.

\begin{table}[htbp]
    \centering
    \begin{tabular}{c|cccc}
    \toprule
        Bit Width & Number of & Encoding Latency  & Energy Consumption & Energy Consumption \\ 
        & Features & ( $\mu s$ ) & ($\mu Ah$) &($\mu J$) \\ 
        \midrule
        2 & 6 & 10 & 0.00016 & 2.11 \\ 
        2 & 81 & 50 & 0.00073 & 9.79 \\
        3 & 6 & 10 & 0.00016 & 2.11 \\ 
        3 & 81 & 70 & 0.00102 & 13.54 \\
        4 & 6 & 10 & 0.00016 & 2.14 \\ 
        4 & 81 & 70 & 0.00112 & 15.62 \\ 
        \bottomrule
    \end{tabular}
    \vspace{0.2cm}
    \caption{Latency and energy requirements for encoding of input features to bit widths 2, 3, and 4 on a micrcontroller with ESP32-S3 CPU.}
    \label{tab:encoding_deployment}
\end{table}
\vspace{-0.4cm}

\textbf{Memory: }The memory overhead of the encoding functionalities including the thresholds amounts to 433 bytes for a 4-bit compression of 81 features or 203 bytes for a 2-bit compression of 6 features. The numbers are given for the source code compiled with Arduino IDE and written to a ESP32-S3 based MCU. For such a device, again, the overhead is negligible as it has 8 MB of Flash memory. But also for smaller devices, e.g. the Arduino Uno that has only 32 KB of flash, this encoding approach can easily be added to most existing approaches.

\subsection{Transmission}
    
We chose to evaluate the data transmission energy and latency requirements in the LTE-M network with the CoAP message protocol, reflecting a widely used setup in the IoT domain. More specifically we used an IoT SIM card from \texttt{q.Beyond}, a self-hosted CoAP server running with the \texttt{aiocoap}\footnote{\url{https://github.com/chrysn/aiocoap}} GitHub repository, and required an acknowledgment for all messages. Notably, measuring network latency is a highly complex task and is dependent on many factors. The numbers presented in Table~\ref{tab:transmission-deployment} only display an example and are not universal. However, the main purpose is to showcase that the possible energy savings when reducing network latency outweigh the cost of simple encoding.

\begin{table}[h!]
    \centering
    \begin{tabular}{c|ccc}
    \toprule
        Byte size & Latency ($ms$) & Energy Consumption ($\mu Ah$) & Energy Consumption ($\mu J$) \\ 
        \midrule
        8 bytes & 21 & 0.6733 & 8968.34 \\ 
        32 bytes & 140 & 4.0779 & 54318.13 \\ 
        128 byte & 250 & 8.4043 & 111945.38 \\ 
        \bottomrule
    \end{tabular}
    \vspace{0.2cm}
    \caption{Energy and latency requirements for sending different amounts of data via CoAP protocol from a microcontroller equipped with ESP32-S3 CPU.}
    \label{tab:transmission-deployment}
\end{table}

\section{Datasets}
\label{app:datasets}
All datasets are downloaded using the API from OpenML \cite{OpenML2020}. 
\begin{table}[H]
    \centering
    \begin{tabular}{lll|m{1.0cm}m{1.0cm}m{1.0cm}m{2.0cm}}
    \toprule
\textbf{Dataset} & \textbf{Short Name} & \textbf{Source} & \textbf{\textit{$N$}}& \textbf{\textit{$K$}}& \textbf{\textit{$\bar{C}$}} & \textbf{OpenML ID} \\\midrule
\californialong & \california & \cite{pace1997sparse} & 20640 & 8 & 8880 & 43979 \\ 
\friedlong & \fried & \cite{friedman1991multivariate, breiman1996bagging} & 40768 & 10 & 1001  & 564 \\ 
\superconductlong & \superconduct & \citep{superconductoriginal}&  21263& 81 & 7449 & 43174 \\ 
\cpulong & \cpu & 
& 8192 & 21 & 2192 & 197  \\ 
\sulfurlong & \sulfur & \cite{fortuna2007soft} & 10081 & 6 & 8636 & 23515 \\ 
\winelong & \wine &\cite{wine_qualitydata} & 6497 & 11 & 241 & 287  \\ \bottomrule
    \end{tabular}
    \vspace{0.2cm}
    \caption{Datasets used where $N$ denotes the size of the dataset, $K$ the number of features and $\bar{C}$ the average number of characteristics (i.e. distinct values) per feature.}
    \label{tab:datasets}
\end{table}

\section{Temperature Schedule Experiments}
\label{app:tau_schedule}

\begin{figure*}[t]
    \centering
    \includegraphics[width=0.7\linewidth]{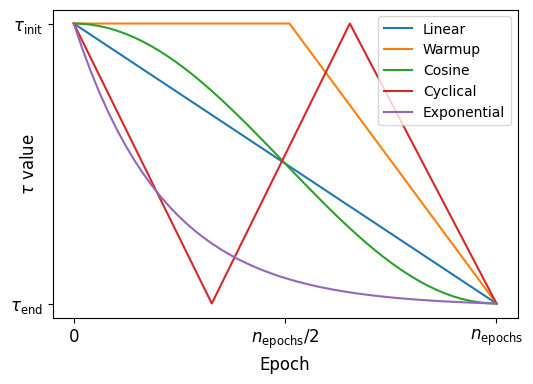}
    \vskip-0.1cm
    \caption{Visualization of different temperature schedules. All schedules start with a temperature value $\tau_{init} = 1$ and end with a temperature value $\tau_{end}$ after $n_{epochs}$ epochs.}
    \label{fig:tau_schedules}
\end{figure*}

We performed experiments to compare different schedules for the temperature parameter $\tau$. The different schedules tested are visualized in Figure \ref{fig:tau_schedules}. 

The performance of Bitwise Soft-Quantization for different schedules was tested with 2 different setups: First, on the wine dataset for $2$ bit quantization and second, on superconduct for $7$ bit quantization. Results are in Table \ref{tab:schedule_results_wine} and Table \ref{tab:schedule_results_superconduct}. In both setups, all temperature schedules were tested on the respective best hyperparameter setup and use an initial $\tau$ value of $\tau_{init}=1$ and the validation MSE of $k=4$-fold cross-validation is reported. Note that we selected these two setups to include each one setting where Bitwise Soft Quantization works well, resp. badly. We also tried higher values for $\tau_{init}$, but that led to worse results. In total, the temperature schedule has an influence, but the exponential schedule together with the options $\tau_{end} = 0.0001$ and $\tau_{end} = 0.001$ that we used in the paper works fine.

\begin{table}[H]
\centering
\begin{tabular}{l|ccc}
\toprule
\multirow{2}{*}{Schedule}& \multicolumn{3}{c}{$\tau_{end}$}   \\
 & 0.0001 & 0.001 & 0.01  \\
\midrule
Linear & 0.6315 & 0.6515 & 0.6421 \\
Warmup & 0.6425 & 0.6456 & 0.6604 \\
Cosine & 0.6401 & 0.6412 & 0.6320 \\
Cyclical & 0.6526 & 0.6473 & 0.6503 \\
Exponential & 0.6300 & \textbf{0.6201} & 0.6412 \\
\bottomrule
\end{tabular}
    \vspace{0.2cm}
\caption{Average validation MSE of Bitwise Soft-Quantization on wine dataset for $2$ bit quantization for different temperature schedules.}
\label{tab:schedule_results_wine}
\end{table}

\begin{table}[H]
\centering
\begin{tabular}{l|ccc}
\toprule
\multirow{2}{*}{Schedule}& \multicolumn{3}{c}{$\tau_{end}$}   \\
 & 0.0001 & 0.001 & 0.01  \\
\midrule
Linear & 0.1615 & 0.1790 & 0.1672 \\
Warmup & 0.1848 & 0.1741 & 0.1737 \\
Cosine & 0.1335 & 0.1336 & 0.1449 \\
Cyclical & 0.1595 & 0.1636 & 0.1896 \\
Exponential & 0.1197 & \textbf{0.1112} & 0.1230 \\
\bottomrule
\end{tabular}
    \vspace{0.2cm}
\caption{Average validation MSE of Bitwise Soft-Quantization on superconduct dataset for $7$ bit quantization for different temperature schedules.}
\label{tab:schedule_results_superconduct}
\end{table}

\section{Hyperparameters}
\label{app:hyperparameter}

The hyperparameter settings used in hyperparameter optimization are listed in Table \ref{tab:hyperparameter}.

\begin{table}[h]
  \centering
  \begin{tabular}{l|ll}
    \toprule
     Name & Values  & Description \\ \midrule
   dropout rate & [0.0, 0.2, 0.4, 0.5] & dropout rate for neurons during training  \\ 
   learning rate & [0.001, 0.0001] & learning rate for optimizer \\ 
   hidden layers & [5, 6, 8, 10] & number of hidden layers \\ 
   \multirow{2}{*}{hidden neurons} & [32, 64, 128, 256, 512, & \multirow{2}{*}{number of neurons per hidden layer} \\
   & 1024, 2048, 4096, 8192]&  \\  
   num epochs & [30, 50, 70] & number of epochs to train \\ 
   decrease factor & [0.001, 0.0001] & factor to decrease $\tau$ during training \\
    \bottomrule
  \end{tabular}
    \vspace{0.2cm}
  \caption{Search Space for Hyperparameter Tuning}
  \label{tab:hyperparameter}
\end{table}

\section{Hardware Setup}
\label{app:sec:hardware}
All experiments were executed on a server equipped with a NVIDIA GeForce RTX\textsuperscript{TM} 4090 GPU with 24GB GDDR6x VRAM in combination with 8 AMD EPYC\textsuperscript{TM} 9124 CPUs. The code was compiled with the GCC toolchain version 12.3.0 in combination with PyTorch 2.1.2 for CUDA 12.1. Choices for the hyperparameter were generated with setting a random seed for the \texttt{random} package in python.

\section{Further Results}
\label{app:result}
\begin{figure*}[!ht]
   \centering
   \includegraphics[width=\textwidth]{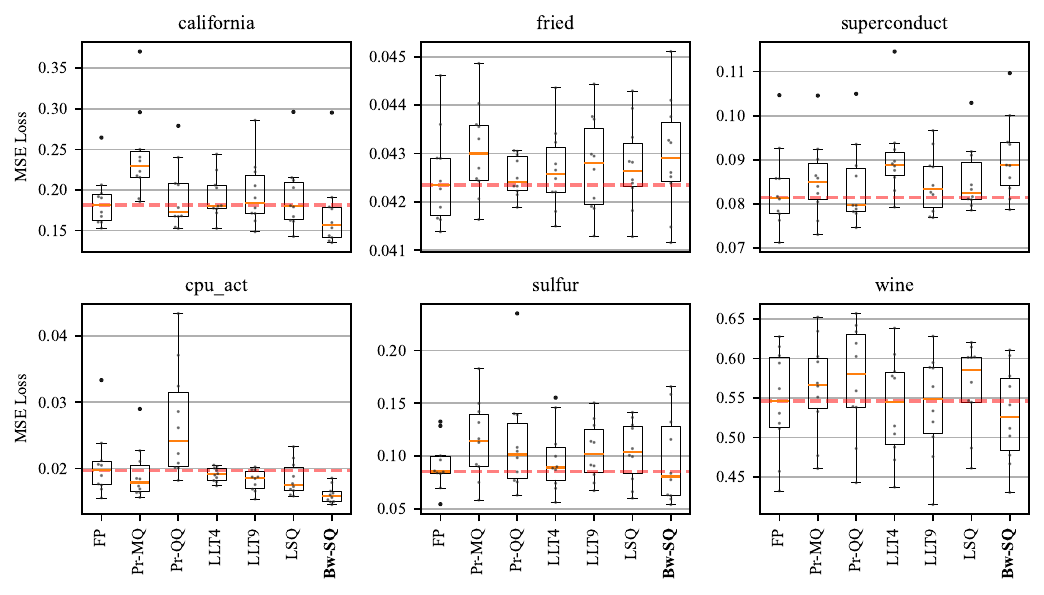}
    \caption{Distribution of the k-fold values for the best average hyperparameter setup per quantization method for a 7-bit quantization level.}
    \label{fig:boxplotkfolds7bits}
 \end{figure*}

\begin{figure*}[!ht]
   \centering
   \includegraphics[width=\textwidth]{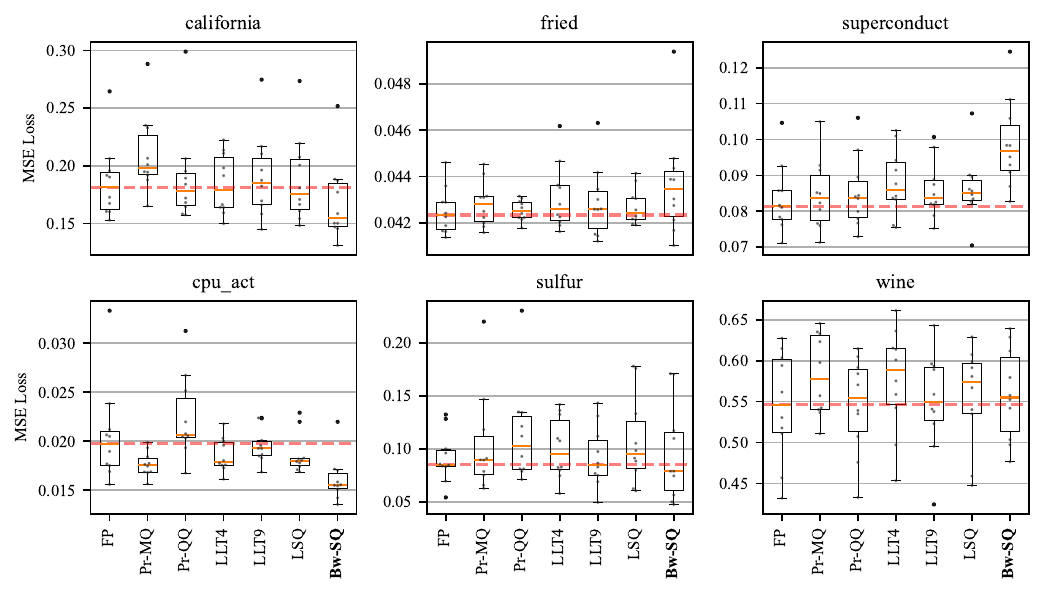}
    \caption{Distribution of the k-fold values for the best average hyperparameter setup per quantization method for a 8-bit quantization level.}
    \label{fig:boxplotkfolds8bits}
 \end{figure*}

The following figures illustrate the distribution of test loss across the individual folds. Note that the scales of the individual datasets also vary across the figures. As the number of bits increases (see Figures \ref{fig:boxplotkfolds7bits} and \ref{fig:boxplotkfolds8bits}), the approaches converge. Furthermore, we add detailed result tables for our experiments and the ablation study. 
\begin{figure*}[!ht]
    \centering
    \includegraphics[width=\linewidth]{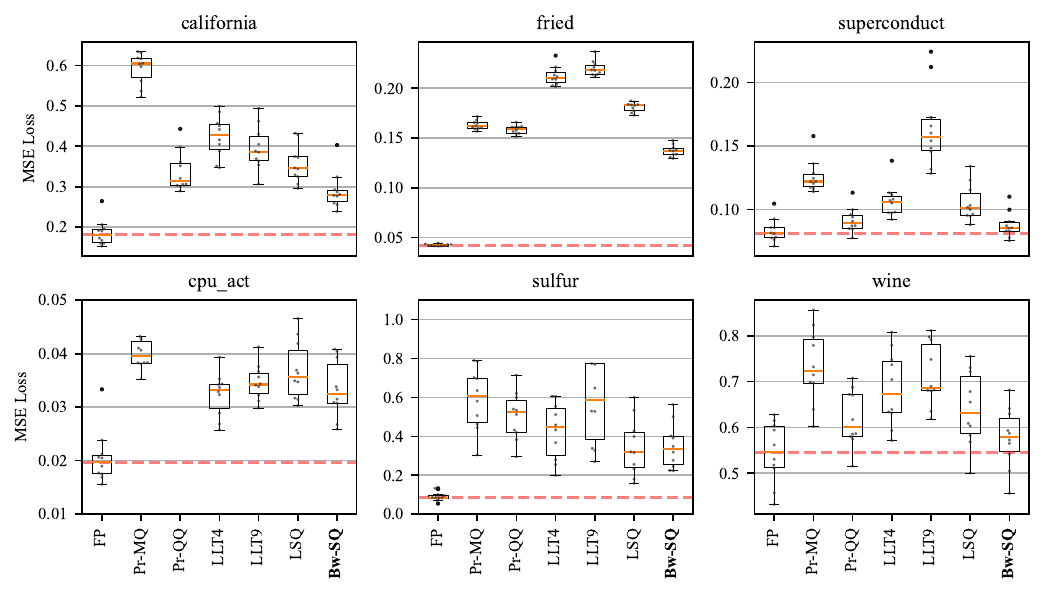}
    \vskip-0.1cm
    \caption{Distribution of the k-fold values for the best average hyperparameter setup per quantization method for a 2-bit quantization level.}
    \label{fig:boxplotkfolds2bits}
\end{figure*}
 \begin{figure*}[!ht]
   \centering
   \includegraphics[width=\textwidth]{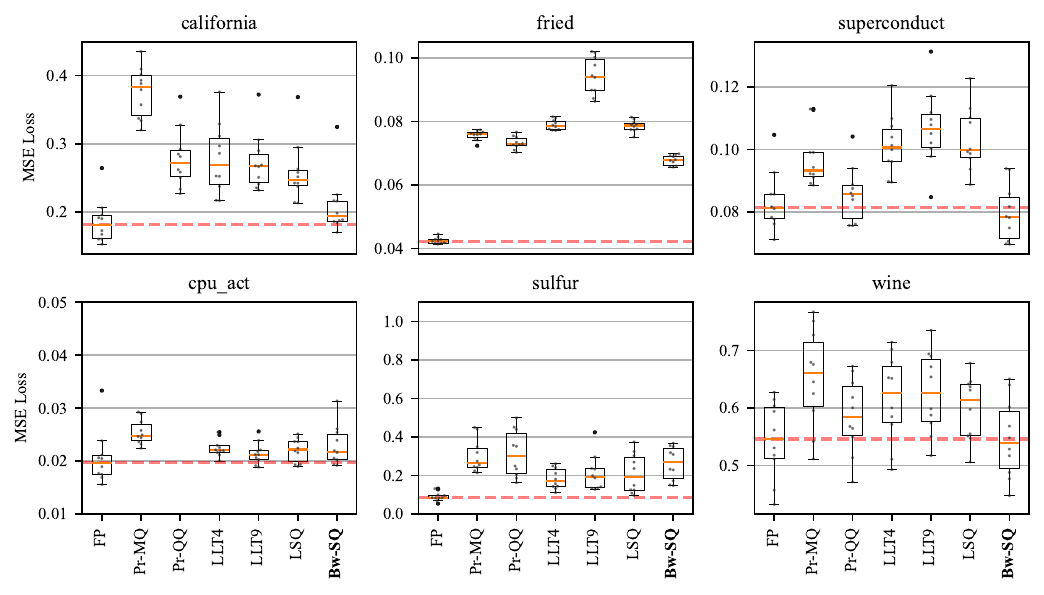}
    \caption{Distribution of the k-fold values for the best average hyperparameter setup per quantization method for a 3-bit quantization level.}
        \label{fig:boxplotkfolds3bits}
 \end{figure*}
 \begin{figure*}[!ht]
   \centering
   \includegraphics[width=\textwidth]{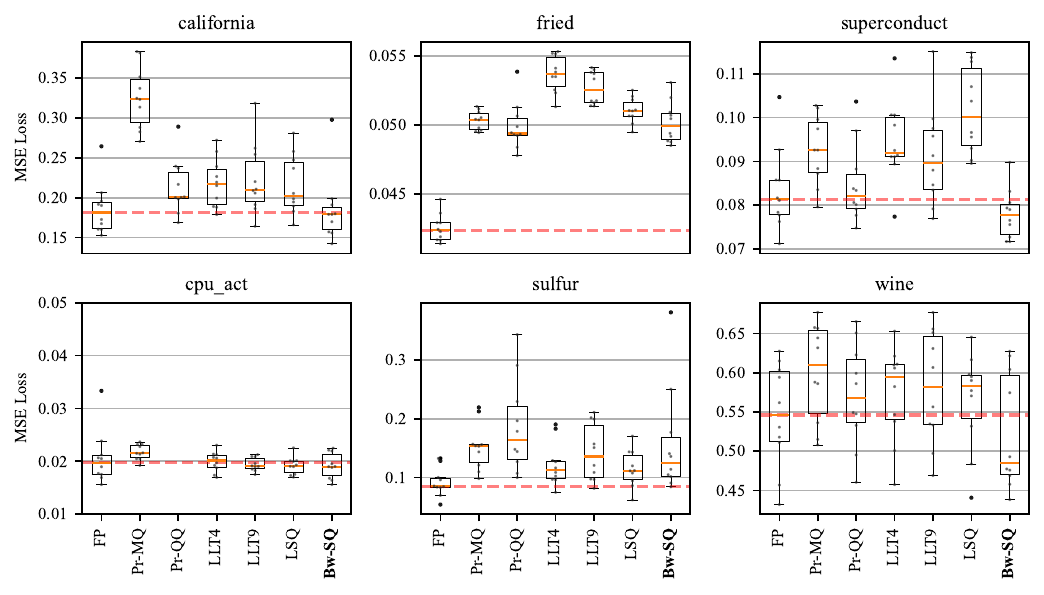}
    \caption{Distribution of the k-fold values for the best average hyperparameter setup per quantization method for a 4-bit quantization level.}
        \label{fig:boxplotkfolds4bits}
 \end{figure*}
 \begin{figure*}[!ht]
   \centering
   \includegraphics[width=\textwidth]{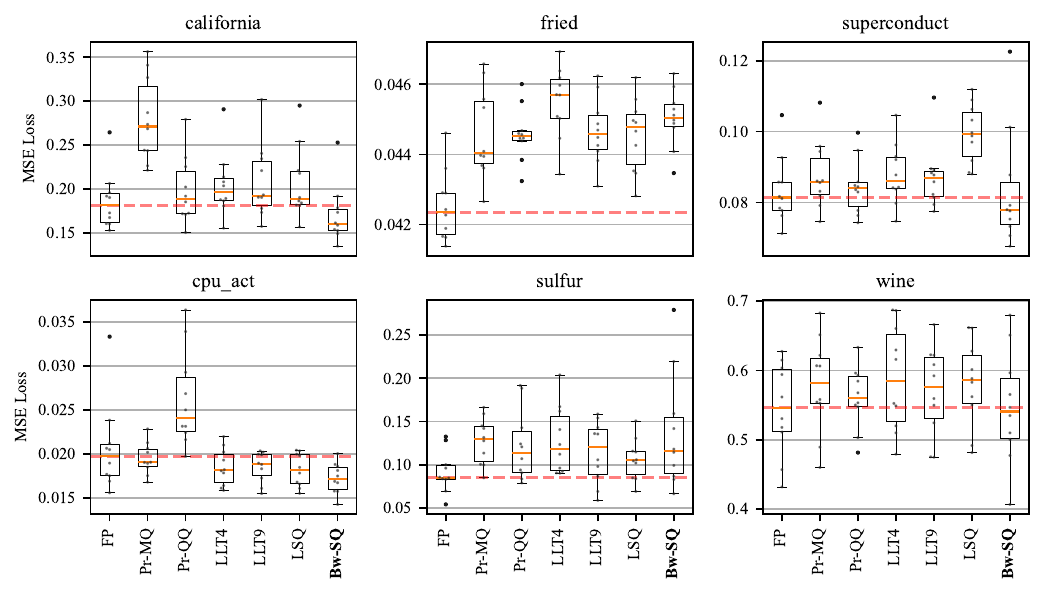}
    \caption{Distribution of the k-fold values for the best average hyperparameter setup per quantization method for a 5-bit quantization level.}
        \label{fig:boxplotkfolds5bits}
 \end{figure*}
 \begin{figure*}[!ht]
   \centering
   \includegraphics[width=\textwidth]{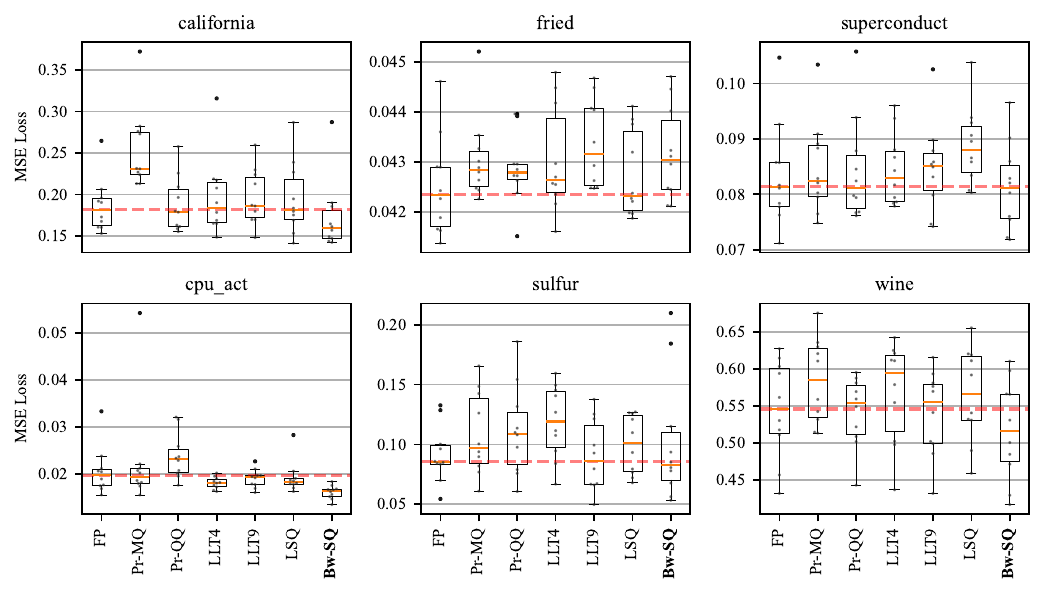}
    \caption{Distribution of the k-fold values for the best average hyperparameter setup per quantization method for a 6-bit quantization level.}
        \label{fig:boxplotkfolds6bits}
 \end{figure*}

\begin{table}[h]
\centering
    \begin{tabular}{lcc|ccccc|c}
\toprule
\multirow{2}{*}{dataset} & \multirow{2}{*}{bits} & \multirow{2}{*}{FP} & \multicolumn{5}{c|}{Comparison models} & \multicolumn{1}{c}{Ours}  \\ &  &  & Pr-MQ & Pr-QQ & LLT4 & LLT9 & LSQ & Bw-SQ \\
\midrule
\multirow{7}{*}{\california} & 2 & 0.186 & 0.593 & 0.338 & 0.424 & 0.396 & 0.356 & \textbf{0.290} \\
 & 3 & 0.186 & 0.376 & 0.279 & 0.278 & 0.273 & 0.259 & \textbf{0.210} \\
 & 4 & 0.186 & 0.326 & 0.213 & 0.219 & 0.222 & 0.216 & \textbf{0.185} \\
 & 5 & 0.186 & 0.279 & 0.199 & 0.204 & 0.208 & 0.207 & \textbf{0.171} \\
 & 6 & 0.186 & 0.254 & 0.189 & 0.197 & 0.197 & 0.195 & \textbf{0.173} \\
 & 7 & 0.186 & 0.242 & 0.192 & 0.192 & 0.196 & 0.192 & \textbf{0.171} \\
 & 8 & 0.186 & 0.210 & 0.189 & 0.184 & 0.191 & 0.188 & \textbf{0.169} \\
\hline
\multirow{7}{*}{\fried} & 2 & 0.043 & 0.163 & 0.159 & 0.212 & 0.220 & 0.181 & \textbf{0.137} \\
 & 3 & 0.043 & 0.076 & 0.073 & 0.079 & 0.094 & 0.079 & \textbf{0.068} \\
 & 4 & 0.043 & \underline{0.050} & \textbf{0.050} & 0.054 & 0.053 & \underline{0.051} & \underline{0.050} \\
 & 5 & 0.043 & \underline{0.045} & \underline{0.045} & \underline{0.045} & \underline{0.045} & \textbf{0.045} & \underline{0.045} \\
 & 6 & 0.043 & \underline{0.043} & \underline{0.043} & \underline{0.043} & \underline{0.043} & \textbf{0.043} & \underline{0.043} \\
 & 7 & 0.043 & \underline{0.043} & \textbf{0.043} & \underline{0.043} & \underline{0.043} & \underline{0.043} & \underline{0.043} \\
 & 8 & 0.043 & \underline{0.043} & \textbf{0.043} & \underline{0.043} & \underline{0.043} & \underline{0.043} & 0.044 \\
\hline
\multirow{7}{*}{\superconduct} & 2 & 0.084 & 0.126 & 0.092 & 0.107 & 0.164 & 0.105 & \textbf{0.088} \\
 & 3 & 0.084 & 0.097 & 0.086 & 0.101 & 0.107 & 0.103 & \textbf{0.080} \\
 & 4 & 0.084 & 0.093 & 0.085 & 0.095 & 0.091 & 0.102 & \textbf{0.078} \\
 & 5 & 0.084 & 0.087 & \underline{0.084} & 0.088 & 0.087 & 0.099 & \textbf{0.083} \\
 & 6 & 0.084 & 0.085 & 0.084 & 0.084 & 0.085 & 0.089 & \textbf{0.081} \\
 & 7 & 0.084 & \underline{0.086} & \textbf{0.084} & 0.090 & \underline{0.085} & \underline{0.086} & 0.090 \\
 & 8 & 0.084 & \textbf{0.085} & \underline{0.085} & 0.088 & \underline{0.086} & \underline{0.086} & 0.099 \\
\hline
\multirow{7}{*}{\cpu} & 2 & 0.021 & 0.043 & 0.316 & \textbf{0.032} & 0.035 & 0.037 & 0.033 \\
 & 3 & 0.021 & 0.025 & 0.197 & 0.022 & \textbf{0.021} & 0.022 & 0.023 \\
 & 4 & 0.021 & 0.022 & 0.120 & 0.020 & \underline{0.019} & \underline{0.019} & \textbf{0.019} \\
 & 5 & 0.021 & 0.019 & 0.026 & 0.019 & 0.019 & 0.018 & \textbf{0.017} \\
 & 6 & 0.021 & 0.023 & 0.024 & 0.018 & 0.019 & 0.019 & \textbf{0.016} \\
 & 7 & 0.021 & 0.019 & 0.027 & 0.019 & 0.018 & 0.018 & \textbf{0.016} \\
 & 8 & 0.021 & 0.018 & 0.022 & 0.019 & 0.020 & 0.019 & \textbf{0.016} \\
\hline
\multirow{7}{*}{\sulfur} & 2 & 0.092 & 0.591 & 0.505 & 0.427 & 1.490 & \textbf{0.343} & \underline{0.350} \\
 & 3 & 0.092 & 0.302 & 0.314 & \textbf{0.184} & 0.212 & 0.212 & 0.264 \\
 & 4 & 0.092 & 0.152 & 0.186 & 0.123 & 0.143 & \textbf{0.115} & 0.158 \\
 & 5 & 0.092 & 0.127 & 0.122 & 0.128 & 0.115 & \textbf{0.106} & 0.137 \\
 & 6 & 0.092 & 0.109 & 0.112 & 0.119 & \textbf{0.091} & 0.100 & 0.102 \\
 & 7 & 0.092 & 0.115 & 0.113 & \underline{0.098} & 0.105 & 0.105 & \textbf{0.097} \\
 & 8 & 0.092 & 0.104 & 0.114 & 0.101 & \textbf{0.093} & 0.108 & 0.096 \\
\hline
\multirow{7}{*}{\wine} & 2 & 0.545 & 0.734 & 0.620 & 0.684 & 0.713 & 0.640 & \textbf{0.577} \\
 & 3 & 0.545 & 0.652 & 0.587 & 0.616 & 0.627 & 0.599 & \textbf{0.547} \\
 & 4 & 0.545 & 0.600 & 0.571 & 0.573 & 0.581 & 0.565 & \textbf{0.524} \\
 & 5 & 0.545 & 0.578 & 0.561 & 0.589 & 0.570 & 0.581 & \textbf{0.547} \\
 & 6 & 0.545 & 0.584 & 0.542 & 0.568 & 0.540 & 0.565 & \textbf{0.518} \\
 & 7 & 0.545 & 0.564 & 0.572 & \underline{0.539} & \underline{0.541} & 0.565 & \textbf{0.528} \\
 & 8 & 0.545 & 0.582 & \textbf{0.546} & 0.576 & \underline{0.550} & \underline{0.556} & \underline{0.559} \\
\bottomrule
\end{tabular}
    \vspace{0.2cm}
\caption{Average MSE for selected hyperparameter setting per dataset and bit width. Minimal values of baseline and our models per row are bold, values within a range of 2.5\% from the minimal value are underlined.}    
\end{table}

\begin{table}[h]
  \resizebox{\textwidth}{!}{
\begin{tabular}{lcc|ccccc|c}
\toprule
\multirow{2}{*}{dataset} & \multirow{2}{*}{bits} & \multirow{2}{*}{FP} & \multicolumn{5}{c|}{Comparison models} & \multicolumn{1}{c}{Ours}  \\ &  &  & Pr-MQ & Pr-QQ & LLT4 & LLT9 & LSQ & Bw-SQ \\
\midrule
\multirow{7}{*}{\california} & 2 & [0.163, 0.210] & [0.565, 0.621] & [0.302, 0.374] & [0.387, 0.462] & [0.357, 0.435] & [0.322, 0.389] & [0.257, 0.323] \\
 & 3 & [0.163, 0.210] & [0.349, 0.402] & [0.248, 0.310] & [0.241, 0.315] & [0.243, 0.303] & [0.226, 0.291] & [0.178, 0.241] \\
 & 4 & [0.163, 0.210] & [0.298, 0.354] & [0.188, 0.238] & [0.196, 0.241] & [0.190, 0.254] & [0.189, 0.243] & [0.154, 0.216] \\
 & 5 & [0.163, 0.210] & [0.245, 0.313] & [0.171, 0.226] & [0.178, 0.230] & [0.178, 0.239] & [0.177, 0.236] & [0.147, 0.194] \\
 & 6 & [0.163, 0.210] & [0.219, 0.289] & [0.164, 0.213] & [0.163, 0.232] & [0.172, 0.221] & [0.163, 0.226] & [0.142, 0.204] \\
 & 7 & [0.163, 0.210] & [0.203, 0.282] & [0.163, 0.222] & [0.172, 0.211] & [0.167, 0.225] & [0.161, 0.223] & [0.137, 0.205] \\
 & 8 & [0.163, 0.210] & [0.185, 0.234] & [0.160, 0.219] & [0.165, 0.203] & [0.164, 0.217] & [0.161, 0.216] & [0.144, 0.194] \\
\hline
\multirow{7}{*}{\fried} & 2 & [0.042, 0.043] & [0.160, 0.166] & [0.156, 0.162] & [0.206, 0.219] & [0.215, 0.225] & [0.178, 0.185] & [0.133, 0.141] \\
 & 3 & [0.042, 0.043] & [0.075, 0.077] & [0.072, 0.075] & [0.078, 0.080] & [0.090, 0.099] & [0.077, 0.080] & [0.067, 0.069] \\
 & 4 & [0.042, 0.043] & [0.050, 0.051] & [0.049, 0.051] & [0.053, 0.055] & [0.052, 0.054] & [0.050, 0.052] & [0.049, 0.051] \\
 & 5 & [0.042, 0.043] & [0.044, 0.046] & [0.044, 0.045] & [0.045, 0.046] & [0.044, 0.045] & [0.044, 0.045] & [0.044, 0.046] \\
 & 6 & [0.042, 0.043] & [0.042, 0.044] & [0.042, 0.043] & [0.042, 0.044] & [0.043, 0.044] & [0.042, 0.043] & [0.042, 0.044] \\
 & 7 & [0.042, 0.043] & [0.042, 0.044] & [0.042, 0.043] & [0.042, 0.043] & [0.042, 0.043] & [0.042, 0.043] & [0.042, 0.044] \\
 & 8 & [0.042, 0.043] & [0.042, 0.044] & [0.042, 0.043] & [0.042, 0.044] & [0.042, 0.044] & [0.042, 0.043] & [0.042, 0.045] \\
\hline
\multirow{7}{*}{\superconduct} & 2 & [0.077, 0.090] & [0.117, 0.135] & [0.085, 0.099] & [0.098, 0.116] & [0.142, 0.187] & [0.095, 0.116] & [0.080, 0.095] \\
 & 3 & [0.077, 0.090] & [0.091, 0.104] & [0.079, 0.092] & [0.095, 0.108] & [0.098, 0.116] & [0.096, 0.111] & [0.073, 0.086] \\
 & 4 & [0.077, 0.090] & [0.087, 0.098] & [0.078, 0.091] & [0.088, 0.101] & [0.083, 0.099] & [0.095, 0.109] & [0.074, 0.082] \\
 & 5 & [0.077, 0.090] & [0.081, 0.094] & [0.079, 0.090] & [0.082, 0.094] & [0.081, 0.094] & [0.093, 0.105] & [0.071, 0.095] \\
 & 6 & [0.077, 0.090] & [0.079, 0.091] & [0.078, 0.091] & [0.080, 0.089] & [0.079, 0.091] & [0.084, 0.094] & [0.076, 0.087] \\
 & 7 & [0.077, 0.090] & [0.079, 0.092] & [0.078, 0.091] & [0.084, 0.097] & [0.080, 0.090] & [0.080, 0.091] & [0.084, 0.097] \\
 & 8 & [0.077, 0.090] & [0.078, 0.092] & [0.078, 0.092] & [0.081, 0.095] & [0.080, 0.092] & [0.080, 0.093] & [0.090, 0.108] \\
\hline
\multirow{7}{*}{\cpu} & 2 & [0.017, 0.024] & [0.035, 0.051] & [0.283, 0.350] & [0.029, 0.035] & [0.032, 0.037] & [0.033, 0.041] & [0.030, 0.037] \\
 & 3 & [0.017, 0.024] & [0.024, 0.027] & [0.154, 0.239] & [0.021, 0.024] & [0.020, 0.023] & [0.020, 0.024] & [0.020, 0.026] \\
 & 4 & [0.017, 0.024] & [0.021, 0.023] & [0.089, 0.151] & [0.019, 0.021] & [0.019, 0.020] & [0.018, 0.020] & [0.017, 0.021] \\
 & 5 & [0.017, 0.024] & [0.018, 0.021] & [0.022, 0.030] & [0.017, 0.020] & [0.017, 0.020] & [0.017, 0.020] & [0.016, 0.018] \\
 & 6 & [0.017, 0.024] & [0.015, 0.031] & [0.020, 0.027] & [0.017, 0.019] & [0.018, 0.020] & [0.017, 0.022] & [0.015, 0.017] \\
 & 7 & [0.017, 0.024] & [0.016, 0.022] & [0.021, 0.033] & [0.018, 0.020] & [0.017, 0.019] & [0.017, 0.020] & [0.015, 0.017] \\
 & 8 & [0.017, 0.024] & [0.017, 0.019] & [0.019, 0.025] & [0.017, 0.020] & [0.018, 0.021] & [0.017, 0.020] & [0.014, 0.018] \\
\hline
\multirow{7}{*}{\sulfur} & 2 & [0.075, 0.109] & [0.476, 0.705] & [0.416, 0.595] & [0.322, 0.532] & [-0.478, 3.457] & [0.238, 0.448] & [0.266, 0.433] \\
 & 3 & [0.075, 0.109] & [0.240, 0.364] & [0.226, 0.401] & [0.145, 0.222] & [0.146, 0.279] & [0.139, 0.285] & [0.202, 0.327] \\
 & 4 & [0.075, 0.109] & [0.124, 0.181] & [0.129, 0.244] & [0.096, 0.149] & [0.108, 0.177] & [0.093, 0.138] & [0.092, 0.224] \\
 & 5 & [0.075, 0.109] & [0.108, 0.146] & [0.093, 0.152] & [0.100, 0.156] & [0.089, 0.140] & [0.089, 0.123] & [0.089, 0.185] \\
 & 6 & [0.075, 0.109] & [0.084, 0.134] & [0.085, 0.139] & [0.097, 0.141] & [0.069, 0.112] & [0.082, 0.118] & [0.064, 0.141] \\
 & 7 & [0.075, 0.109] & [0.088, 0.142] & [0.077, 0.148] & [0.075, 0.121] & [0.085, 0.125] & [0.084, 0.125] & [0.067, 0.128] \\
 & 8 & [0.075, 0.109] & [0.070, 0.138] & [0.080, 0.148] & [0.080, 0.122] & [0.072, 0.113] & [0.077, 0.138] & [0.063, 0.128] \\
\hline
\multirow{7}{*}{\wine} & 2 & [0.497, 0.593] & [0.676, 0.791] & [0.575, 0.664] & [0.626, 0.743] & [0.664, 0.763] & [0.582, 0.699] & [0.530, 0.625] \\
 & 3 & [0.497, 0.593] & [0.591, 0.713] & [0.540, 0.634] & [0.561, 0.671] & [0.576, 0.679] & [0.559, 0.639] & [0.498, 0.596] \\
 & 4 & [0.497, 0.593] & [0.555, 0.645] & [0.524, 0.618] & [0.529, 0.616] & [0.529, 0.633] & [0.520, 0.610] & [0.471, 0.577] \\
 & 5 & [0.497, 0.593] & [0.529, 0.628] & [0.529, 0.593] & [0.534, 0.644] & [0.524, 0.616] & [0.536, 0.625] & [0.489, 0.605] \\
 & 6 & [0.497, 0.593] & [0.542, 0.625] & [0.508, 0.576] & [0.519, 0.617] & [0.498, 0.581] & [0.520, 0.611] & [0.469, 0.566] \\
 & 7 & [0.497, 0.593] & [0.519, 0.608] & [0.521, 0.623] & [0.493, 0.586] & [0.494, 0.587] & [0.525, 0.604] & [0.485, 0.572] \\
 & 8 & [0.497, 0.593] & [0.546, 0.618] & [0.503, 0.588] & [0.530, 0.622] & [0.506, 0.594] & [0.512, 0.599] & [0.519, 0.599] \\
\bottomrule
\end{tabular}}
    \vspace{0.2cm}
\caption{95\%-confidence intervals of $10$-fold results for selected hyperparameter settings per dataset and bit width.}
\end{table}

\begin{table}[h]
\centering
    \begin{tabular}{lc|ccc|c}
\toprule
\multirow{2}{*}{dataset} & \multirow{2}{*}{bits} & \multicolumn{3}{c|}{Comparison models} & \multicolumn{1}{c}{Ours}  \\ &  & SQ & Bw-MQ & Bw-QQ & Bw-SQ \\
\midrule
\multirow{7}{*}{\california} & 2 & 0.288 & 0.595 & 0.329 & \textbf{0.290} \\
 & 3 & 0.217 & 0.377 & 0.265 & \textbf{0.210} \\
 & 4 & 0.214 & 0.312 & 0.207 & \textbf{0.185} \\
 & 5 & 0.208 & 0.262 & 0.182 & \textbf{0.171} \\
 & 6 & 0.200 & 0.230 & \underline{0.177} & \textbf{0.173} \\
 & 7 & 0.202 & 0.199 & \textbf{0.171} & \underline{0.171} \\
 & 8 & 0.205 & 0.180 & 0.173 & \textbf{0.169} \\
\hline
\multirow{7}{*}{\fried} & 2 & 0.137 & 0.164 & 0.158 & \textbf{0.137} \\
 & 3 & 0.070 & 0.074 & 0.073 & \textbf{0.068} \\
 & 4 & 0.052 & \underline{0.050} & \underline{0.050} & \textbf{0.050} \\
 & 5 & 0.048 & \underline{0.045} & \underline{0.045} & \textbf{0.045} \\
 & 6 & 0.046 & \underline{0.044} & \textbf{0.043} & \underline{0.043} \\
 & 7 & 0.047 & \underline{0.044} & 0.045 & \textbf{0.043} \\
 & 8 & 0.047 & \textbf{0.043} & 0.045 & \underline{0.044} \\
\hline
\multirow{7}{*}{\superconduct} & 2 & 0.094 & 0.134 & \underline{0.090} & \textbf{0.088} \\
 & 3 & 0.088 & 0.101 & 0.083 & \textbf{0.080} \\
 & 4 & 0.087 & 0.092 & \underline{0.079} & \textbf{0.078} \\
 & 5 & 0.097 & 0.093 & 0.086 & \textbf{0.083} \\
 & 6 & 0.094 & 0.089 & \underline{0.082} & \textbf{0.081} \\
 & 7 & 0.089 & 0.097 & \textbf{0.084} & 0.090 \\
 & 8 & 0.085 & \underline{0.092} & \textbf{0.090} & 0.099 \\
\hline
\multirow{7}{*}{\cpu} & 2 & 0.034 & 0.035 & 0.160 & \textbf{0.033} \\
 & 3 & 0.027 & \textbf{0.022} & 0.037 & 0.023 \\
 & 4 & 0.025 & 0.020 & 0.025 & \textbf{0.019} \\
 & 5 & 0.023 & 0.020 & 0.023 & \textbf{0.017} \\
 & 6 & 0.025 & 0.017 & 0.019 & \textbf{0.016} \\
 & 7 & 0.024 & 0.017 & 0.018 & \textbf{0.016} \\
 & 8 & 0.025 & 0.018 & 0.019 & \textbf{0.016} \\
\hline
\multirow{7}{*}{\sulfur} & 2 & 0.348 & 0.593 & 0.498 & \textbf{0.350} \\
 & 3 & 0.305 & 0.287 & 0.374 & \textbf{0.264} \\
 & 4 & 0.175 & 0.184 & 0.210 & \textbf{0.158} \\
 & 5 & 0.152 & \textbf{0.120} & 0.163 & 0.137 \\
 & 6 & 0.129 & \textbf{0.099} & 0.126 & 0.102 \\
 & 7 & 0.134 & \textbf{0.087} & 0.123 & 0.097 \\
 & 8 & 0.123 & \textbf{0.090} & 0.114 & 0.096 \\
\hline
\multirow{7}{*}{\wine} & 2 & 0.610 & 0.733 & 0.596 & \textbf{0.577} \\
 & 3 & 0.584 & 0.645 & 0.560 & \textbf{0.547} \\
 & 4 & 0.567 & 0.582 & \textbf{0.514} & \underline{0.524} \\
 & 5 & 0.583 & \underline{0.519} & \textbf{0.508} & 0.547 \\
 & 6 & 0.585 & \textbf{0.496} & 0.509 & 0.518 \\
 & 7 & 0.566 & \textbf{0.492} & \underline{0.500} & 0.528 \\
 & 8 & 0.584 & \textbf{0.517} & 0.561 & 0.559 \\
\bottomrule
\end{tabular}
    \vspace{0.2cm}
\caption{Ablation Study: Average MSE for selected hyperparameter setting per dataset and bit width. Minimal values are bold, values within a range of 2.5\% from the minimal value are underlined.}   
\label{tab:ablation_full}
\end{table}

\begin{table}
\end{table}

\end{document}